# Full end-to-end diagnostic workflow automation of 3D OCT via foundation model-driven AI for retinal diseases

Jinze Zhang[#,1], Jian Zhong[#,2], Li Lin[#,2,3], Jiaxiong Li[#,1], Ke Ma[4], Naiyang Li[5,6], Meng Li[7], Yuan Pan[1], Zeyu Meng[1], Mengyun Zhou[1], Shang Huang[1], Shilong Yu[1], Zhengyu Duan[1], Sutong Li[5,6], Honghui Xia[8], Juping Liu[9], Dan Liang[1], Yantao Wei[1], Xiaoying Tang[*,2,10], Jin Yuan[*,11] and Peng Xiao[*,1]

[1]State Key Laboratory of Ophthalmology, Zhongshan Ophthalmic Center, Sun Yat-sen University, Guangdong Provincial Key Laboratory of Ophthalmology and Visual Science, Guangzhou, China
[2]Department of Electrical and Electronic Engineering, Southern University of Science and Technology, Shenzhen, China
[3]Department of Electrical and Electronic Engineering, The University of Hong Kong, Hong Kong SAR, China
[4]Department of Health Technology and Informatics, The Hong Kong Polytechnic University, Hong Kong, China
[5]Eye Center, Zhongshan City People's Hospital, Zhongshan, Guangdong, China
[6]Guangdong Medical University, Zhanjiang, Guangdong, China
[7]Department of Ophthalmology, Department of Ophthalmology，The Third Affiliated Hospital of Sun Yat-sen University, Guangzhou, China
[8]Department of Ophthalmology, Zhaoqing Gaoyao People's Hospital, Zhaoqing, China
[9]Tianjin Key Laboratory of Retinal Functions and Diseases, Tianjin Branch of National Clinical Research Center for Ocular Disease, Eye Institute and School of Optometry, Tianjin Medical University Eye Hospital, Tianjin, China
[10]Jiaxing Research Institute, Southern University of Science and Technology, Jiaxing, China
[11]Beijing Tongren Eye Center, Beijing Tongren Hospital, Capital Medical University, Beijing Key Laboratory of Ophthalmology & Visual Sciences, Beijing, China

These authors contributed equally: Jinze Zhang, Jian Zhong, Li Lin, Jiaxiong Li

Corresponding authors: Xiaoying Tang (tangxy@sustech.edu.cn), Jin Yuan (yuanjincornea@126.com) and Peng Xiao (xiaopengaddis@hotmail.com)

**Abstract**

Optical coherence tomography (OCT) has revolutionized retinal disease diagnosis with its high-resolution and three-dimensional imaging nature, yet its full diagnostic automation in clinical practices remains constrained by multi-stage workflows and conventional single-slice single-task AI models. We present Full-process OCT-based Clinical Utility System (FOCUS), a foundation model-driven framework enabling end-to-end automation of 3D OCT retinal disease diagnosis. FOCUS sequentially performs image quality assessment with EfficientNetV2-S, followed by abnormality detection and multi-disease classification using a fine-tuned Vision Foundation Model. Crucially, FOCUS leverages a unified adaptive aggregation method to intelligently integrate 2D slices-level predictions into comprehensive 3D patient-level diagnosis. Trained and tested on 3,300 patients (40,672 slices), and externally validated on 1,345 patients (18,498 slices) across four different-tier centers and diverse OCT devices, FOCUS achieved high F1 scores for quality assessment (99.01%), abnormally detection (97.46%), and patient-level diagnosis (94.39%). Real-world validation across centers also showed stable performance (F1: 90.22%-95.24%). In human-machine comparisons, FOCUS matched expert performance in abnormality detection (F1: 95.47% vs 90.91%) and multi-disease diagnosis (F1: 93.49% vs 91.35%), while demonstrating better efficiency. FOCUS automates the image-to-diagnosis pipeline, representing a critical advance towards unmanned ophthalmology with a validated blueprint for autonomous screening to enhance population scale retinal care accessibility and efficiency.

## Introduction

Optical coherence tomography (OCT)[1,2] has become a cornerstone of modern ophthalmology, revolutionizing retinal disease management with its high-resolution, 3D imaging capabilities[3,4]. However, its transformative potential for large-scale screening remains unrealized due to its complex and labor-intensive manual diagnostic workflow[5,6]. Accurate OCT diagnosis demands a sequence of expert-dependent steps: meticulous image acquisition, targeted selection of pathology-relevant slices, and expert interpretation to guide clinical assessment[7]. These demands restrict the use of OCT for large-scale screening, particularly in primary care settings[8], where the lack of expert interpretation thus compromises diagnostic accuracy. In response, artificial intelligence (AI) has emerged as a powerful solution[9–11], with its rapid evolution in ophthalmic imaging representing a pivotal step towards the future of intelligent and increasingly automated healthcare.

Despite this promise, the first wave of OCT-AI tools has proven insufficient to automate the full clinical journey. Most critically, they are designed for isolated tasks, such as quality classification[12–14] or disease detection[15–18], failing to replicate the integrated, end-to-end nature of expert reasoning. Prior attempts to bridge this gap, such as the semi-automated segmentation pipelines by De Fauw et al.[19], represented significant progress in linking diagnostic steps. However, these systems often remain dependent on high-quality, standardized inputs or explicit segmentation priors, lacking the resilience to handle raw, uncrated data streams. This limitation restricts their utility for large-scale screening and fully automated OCT deployment and home-based medical care[20].

Alongside this workflow limitation, a fundamental methodological challenge persists: retinal pathologies are inherently three-dimensional[21]. Current automated diagnostic strategies predominantly fall into two paradigms: slice-level 2D analysis and volumetric 3D modeling. While 2D approaches are computationally efficient, they inherently treat slices in isolation[17,22,23]. This limitation not only compromises diagnostic accuracy by missing crucial volumetric context but also fails to automate the laborious task of manual scan selection[7]. Conversely, researchers have developed volumetric AI approaches to capture this 3D context directly, ranging from 3D-CNNs to recent 3D Transformer architectures[24–27]. Although these models theoretically utilize full spatial information, they often incur prohibitive computational costs and require massive volumetric annotations. Furthermore, they frequently struggle with the significant data heterogeneity of real-world clinical settings.

Furthermore, although foundation models such as RETFound[28] and VisionFM[29] have significantly advanced generalizability via massive pre-training, their inherent 2D architecture creates a barrier[30]. Directly extending them to 3D diagnosis requires bridging 2D feature extraction with volumetric interpretation. While Multiple Instance Learning (MIL) offers a pathway to synthesize these features, classical MIL approaches relying on rigid aggregation often dilute sparse, 'needle-in-a-haystack' pathological signals[31–34]. As a result, existing volumetric frameworks cannot effectively capitalize on the generalization benefits provided by these foundation models, limiting their robustness in diverse clinical settings. These limitations highlight the urgent need for a unified OCT-AI system that can seamlessly integrate the robustness of foundation models with the volumetric context required for clinical precision, thereby faithfully mirroring real-world diagnostic workflows.

To bridge the critical gap between current AI prototypes and real-world ophthalmic practice, we introduce FOCUS (Full-process OCT-based Clinical Utility System), an end-to-end AI framework engineered to automate the complete clinical OCT workflow. Distinct from prevalent systems that optimize isolated tasks on curated datasets, FOCUS is architected to mimic the comprehensive clinical pipeline, seamlessly integrating quality control, anomaly triage, and volumetric diagnosis into a unified workflow. Technically, the system employs a workflow-adaptive architecture that leverages a Vision Foundation Model enhanced by a Prompt Decoder and a Unified Adaptive Aggregation Classifier (UAAC). The MIL paradigm within FOCUS enables the robust synthesis of patient-level diagnoses from 2D slice streams, effectively capturing sparse pathological features within large volumes without the computational constraints of 3D-CNNs. By automating the full diagnostic chain, FOCUS overcomes barriers related to data heterogeneity and workflow fragmentation, offering a scalable solution for standardized population-level screening (Figure 1).

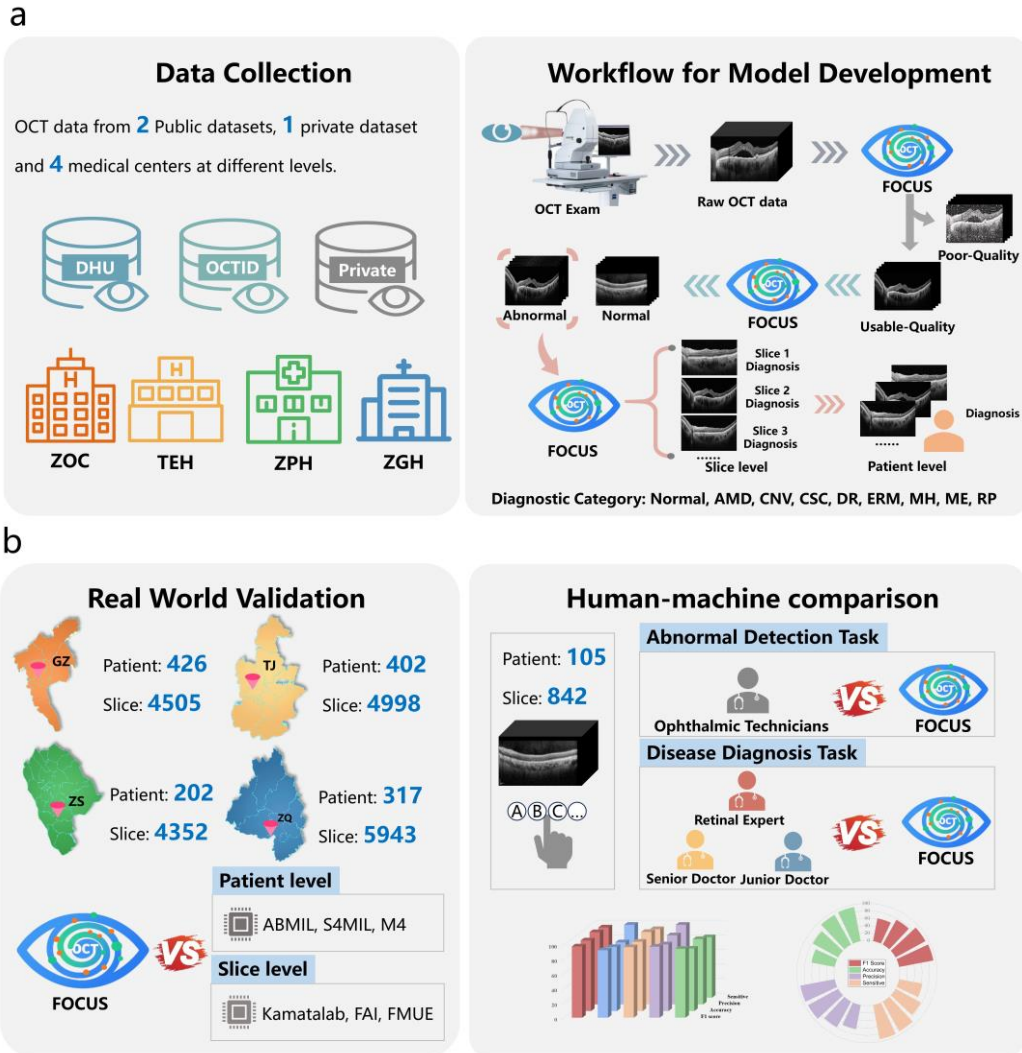

**Figure 1.** Overview and workflow of FOCUS. a. The development of FOCUS. OCT data from 2 public datasets and 4 medical centers at different levels were collected for FOCUS's quality assessment, abnormality detection and multi-level disease diagnosis models. b. Real-world validation and human-machine comparison experiment. We verified the performance of FOCUS in the real world and compared it with different comparable methods and different levels of doctors.

Abbreviations: AMD, age-related macular degeneration; CNV, choroidal neovascularization; CSC, central serous chorioretinopathy; DR, diabetic retinopathy; MH, macular hole; ME, macular edema; ERM, epiretinal membranes; RP, retinitis pigmentosa; OCTID, Optical Coherence Tomography Image Database; ZOC, Zhongshan Ophthalmic Center; THE, Tianjin Medical University Eye Hospital; ZPH, Zhongshan People's Hospital; ZGH, Zhaoqing Gaoyao People's Hospital; GZ, Guangzhou City; TJ: Tianjin City; ZS: Zhongshan City; ZQ: Zhaoqing City.

## Results

### Quality Classification and Abnormal Detection Task

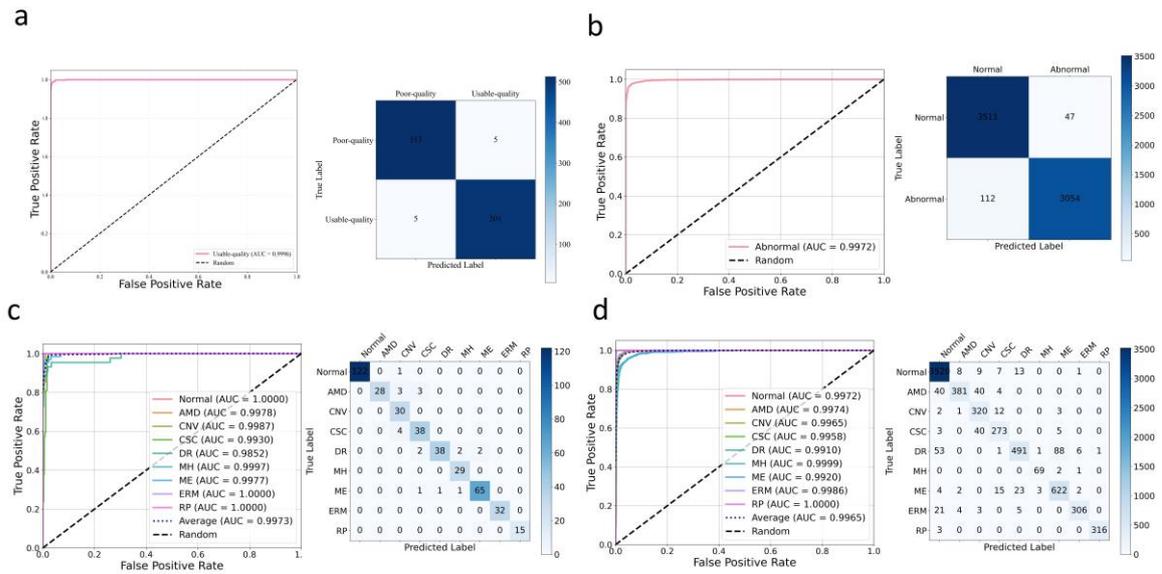

**Figure 2.** Performance of FOCUS of internal dataset. a. Receiver operating characteristic (ROC) curves and confusion matrices of quality control task. b. ROC and confusion matrices of the abnormal detection task. c. ROC and confusion matrices of the abnormal detection task of the patient-level multi-disease diagnosing task. d. ROC and confusion matrices of the abnormal detection task of the slice-level multi-disease diagnosing task.

## Multi-Level Multi-Disease Diagnosing Task

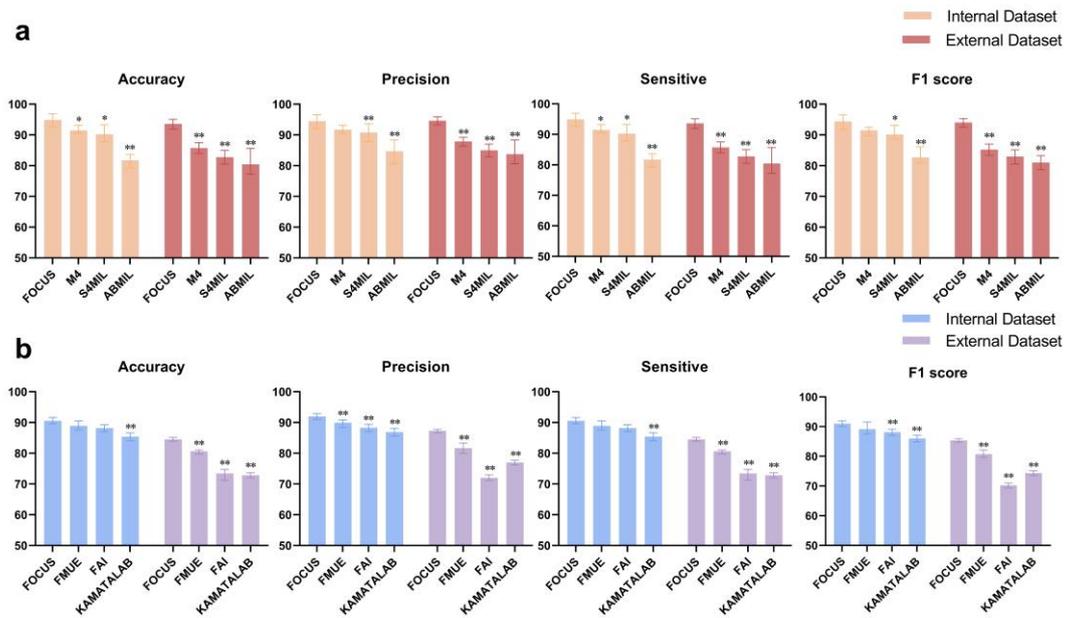

**Figure 3.** Comparison of FOCUS with other models. a. Patient-level performance comparison. (B) Slice-level performance comparison. *p < 0.05 and **p < 0.005

indicate significant differences compared with FOCUS.

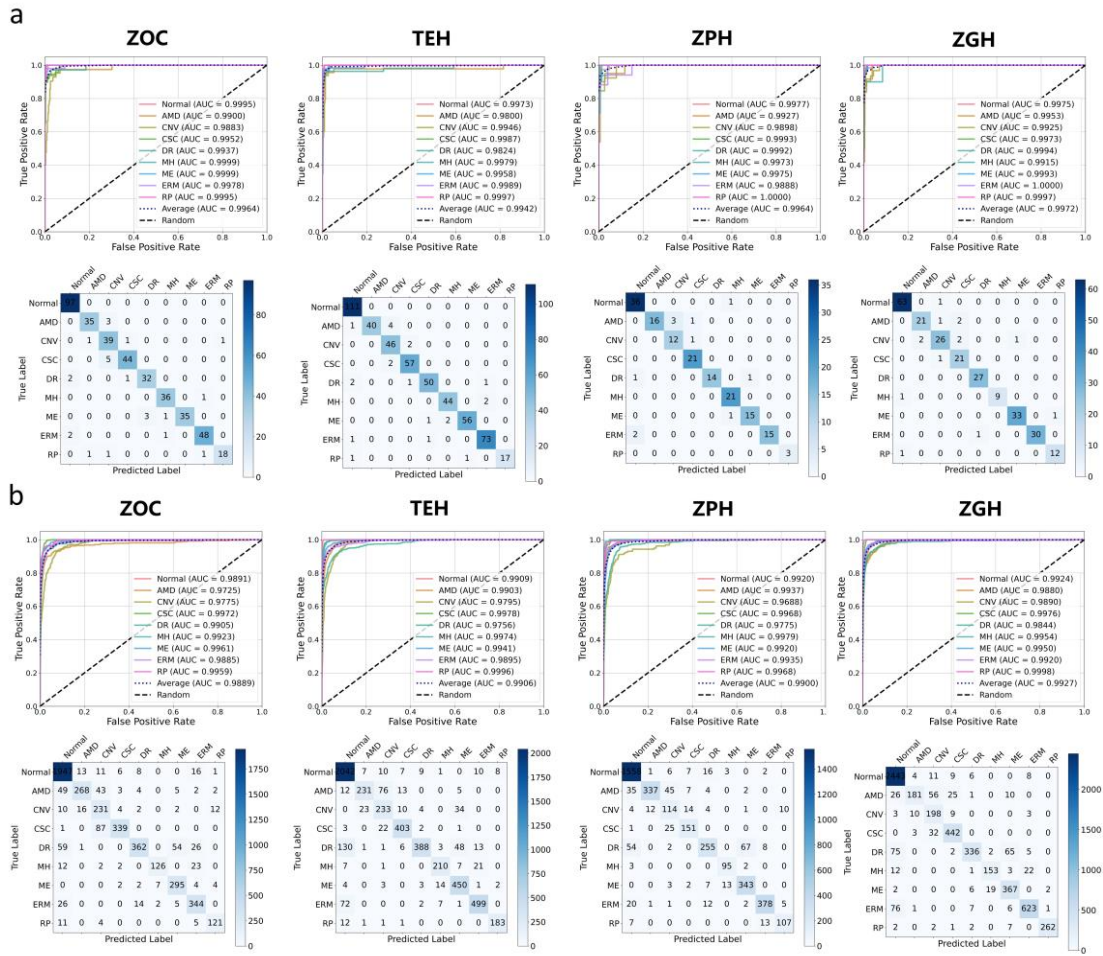

**Figure 4**. Performance of the multi-level multi-disease diagnosing task of each external center. a. Patient-level multi-disease diagnosing receiver operating characteristic (ROC) curves and confusion matrices of the external dataset. b. Slice-level multi-disease diagnosing ROC curves and confusion matrices of the external dataset.

Abbreviations: AMD, age-related macular degeneration; CNV, choroidal neovascularization; CSC, central serous chorioretinopathy; DR, diabetic retinopathy; MH, macular hole; ME, macular edema; ERM, epiretinal membranes; RP, retinitis pigmentosa; OCTID, Optical Coherence Tomography Image Database; ZOC, Zhongshan Ophthalmic Center; THE, Tianjin Medical University Eye Hospital; ZPH, Zhongshan People's Hospital; ZGH, Zhaoqing Gaoyao People's Hospital; GZ, Guangzhou City; TJ: Tianjin City; ZS: Zhongshan City; ZQ: Zhaoqing City.

## Human-Machine Comparison

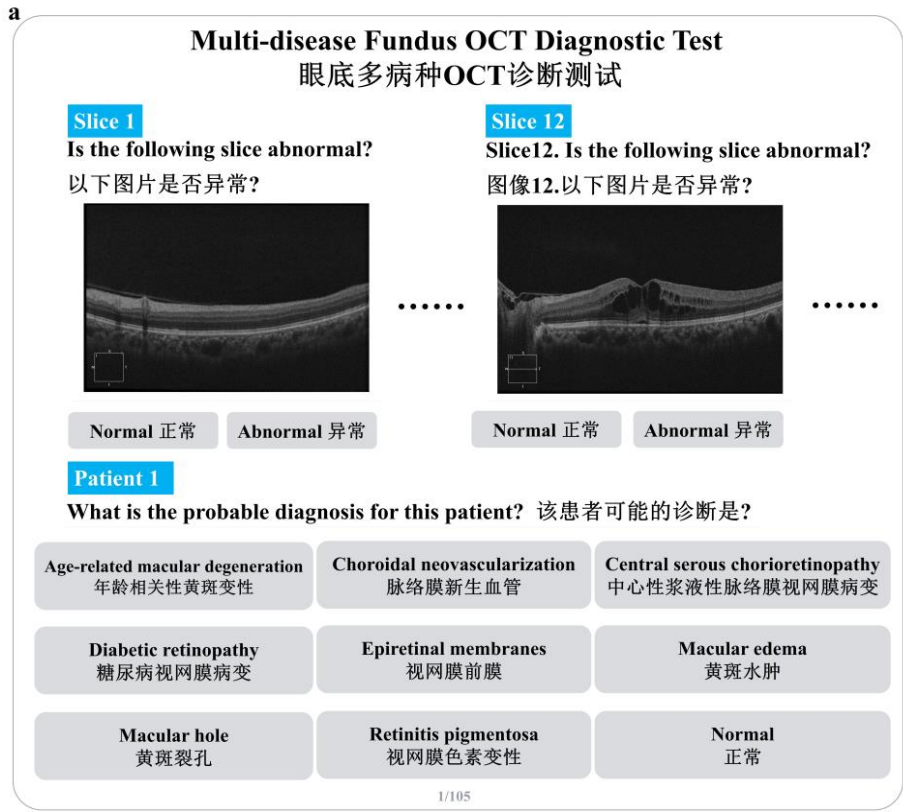

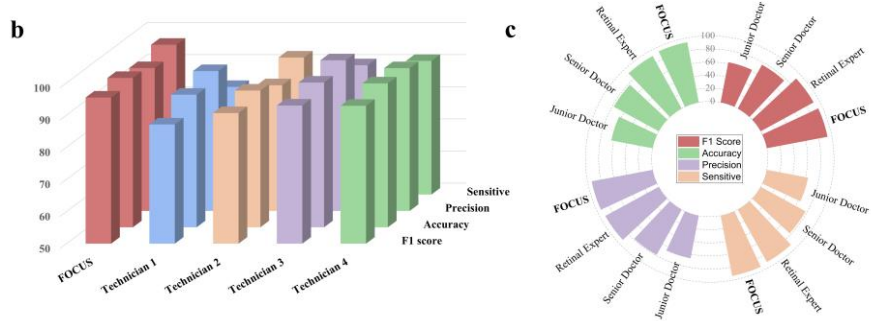

Figure 5. Human-machine comparison a. Online retinal OCT image interpretation system, comprising both image anomaly detection and disease diagnosis modules. b. Performance comparison between FOCUS and four ophthalmic technicians in anomaly detection tasks. c. Diagnostic capability comparison between FOCUS and nine clinicians with varying experience levels in disease diagnosis tasks.

## Discussion

In this study, we developed and validated FOCUS, a system designed to overcome the critical limitations of prior OCT-AI by successfully automating the full diagnostic workflow. Our end-to-end system operates as a seamless, sequential pipeline that mirrors clinical reasoning, progressing from autonomous image quality assessment to abnormality detection and final classification. Most critically, it advances beyond prior models by integrating fragmented slice-level findings into a cohesive, patient-level diagnosis, thereby aligning directly with real-world clinical practice. Our findings demonstrate that this integrated approach achieves high diagnostic accuracy and robust generalizability, validated across a diverse, real-world dataset from multiple centers and devices. To our knowledge, FOCUS is the first OCT-AI system to successfully translate the multi-step, expert-level diagnostic process into a deployable, end-to-end solution. This work therefore represents a critical shift from developing fragmented models to engineering a clinically-aligned, deployable tool capable of true workflow automation.

The landscape of OCT-AI is evolving from narrow, single-disease models to more comprehensive, multi-disease systems. The advent of ophthalmic foundation models like RETFound[28], VisionFM[29], and EyeCLIP[36] has been pivotal in this shift. Trained on massive and diverse datasets, these models possess enhanced robustness and an inherent capability for multi-disease classification that surpasses traditional architectures. However, this power is not a turn-key solution. These foundational models often exhibit moderate out-of-the-box accuracy in specific diagnostic tasks[37], requiring significant fine-tuning to reach clinical-grade performance. For instance, Peng et al. achieved high accuracy only after adapting RETFound with specialized uncertainty analysis methods[17]. It is precisely this state-of-the-art principle that we applied to FOCUS. By building upon a fine-tuned Vision FM[29] as well as being trained by multi-center, multi-device data, we successfully harnessed its robust, multi-disease capabilities to meet the high-accuracy demands of a real-world clinical environment. Despite these advances in multi-disease recognition, a critical limitation persists: the vast majority of these models still operate on isolated 2D slices. This not only prevents the full utilization of rich 3D volumetric context but also fails to address the clinical need for a unified, patient-level diagnosis. While early attempts to use 3D architectures showed promise in referral recommendations[19], they were often computationally intensive and limited in their diagnostic scope. Technically, FOCUS establishes a MIL framework to process volumetric data efficiently. Instead of employing computationally intensive 3D convolutional kernels, the system decouples feature encoding from volumetric aggregation: it utilizes a fine-tuned Vision Foundation Model to extract semantic representations from individual 2D B-scans, which are then dynamically fused by a UAAC. This hierarchical design enables precise, automated patient-level diagnoses directly from raw slice streams. The UAAC empowers the model to dynamically calibrate the influence of each B-scan based on its clinical diagnostic contribution and associated prediction uncertainty. By distinguishing between high-confidence pathological evidence and ambiguous findings, this module ensures that the final patient-level diagnosis is derived from the most reliable slice-level cues. This selective aggregation mechanism maintains robust performance even in challenging scenarios where definitive pathological signs are sparse or interspersed with equivocal features. This advancement not only bridges the gap between research prototypes and clinical deployment but also lays the foundation for scalable, population-level screening and standardized retinal disease management.

A core principle of FOCUS's design is to emulate the real-world, multi-stage ophthalmic workflow[5,6,38], where different tasks are handled by professionals with varying expertise. To validate our system in a clinically meaningful way, we therefore designed a human-machine comparison that mirrored this tiered structure. We benchmarked the abnormality detection module against ophthalmic technicians and the multi-disease diagnostic module against retinal specialists. Across the comparison, FOCUS demonstrated performance that was either comparable to or statistically superior to its human counterparts. This robust, expert-level performance highlights its potential to standardize quality and assist clinicians across the entire diagnostic pathway. Notably, in the human–AI comparison, ophthalmologists showed considerable variability when distinguishing AMD from CNV, as they were restricted to OCT imaging alone. This reflects the inherent difficulty of making this distinction based solely on OCT B-scans, since clinicians typically integrate multimodal information such as angiography, fundus findings and clinical history to reach a definitive diagnosis. In contrast, FOCUS leverages large-scale pretraining to capture subtle structural differences that may be difficult for humans to consistently perceive, and integrates information across the entire 3D volume through slice-level aggregation. These design features enable the system to generate more consistent decisions when operating under imaging-only conditions.

Despite its promising performance, our study has several limitations. First, the training and validation datasets were mainly collected from Chinese medical centers. This demographic concentration may limit the epidemiological generalizability of FOCUS to populations with different ethnic backgrounds, disease spectra, or clinical workflows. To partially mitigate this limitation, our multicenter cohort was intentionally constructed to include data from geographically and socioeconomically diverse regions, including a northern metropolitan center (Tianjin), southern metropolitan centers (Guangzhou and Zhongshan), and a less developed region (Zhaoqing). Nevertheless, broader validation in international populations and across additional device vendors will be needed to further enhance generalizability. Second, the retrospective nature of our evaluation represents a key limitation. The validation datasets were collected from ophthalmic departments in tertiary and general hospitals, where the prevalence of retinal pathologies is significantly higher than in the general population. Consequently, the reported performance reflects the system's utility as a diagnostic aid in clinical settings but may not fully predict its behavior in mass screening scenarios where disease incidence is low. Prospective studies conducted in diverse live settings are essential to definitively validate the system's performance under natural disease distributions, evaluate its impact on workflow efficiency, and assess user acceptance among clinicians. Furthermore, the current iteration of FOCUS has not yet integrated recent advances in large language models[39–42]. Such models offer the potential to support interactive dialogue, automate report generation, and provide more interpretable decision-making processes. Future versions could leverage this technology to enable interactive diagnostic dialogues, automate the generation of clinical reports, and provide even more transparent, human-readable justifications for its findings[43,44]. This would not only enhance user trust but also further embed the system into the collaborative fabric of clinical care. However, it is also important to note that integrating such systems into medical diagnostics would bring significant regulatory challenges. This includes validating the generated narratives, preventing hallucinations, and complying with the regulations of medical devices in specific jurisdictions, future work must carefully evaluate these regulatory considerations. Finally, the transition from a research prototype to a widely deployed clinical tool necessitates rigorous adherence to data security and ethical standards. While this study relied on strictly desensitized retrospective data to ensure privacy, real-world implementation demands robust privacy-preserving infrastructures. To fully comply with evolving data protection regulations, future iterations of FOCUS must integrate advanced security measures—including end-to-end encryption and strict access control protocols. Addressing these ethical and legal imperatives is as crucial as diagnostic accuracy for the successful adoption of autonomous diagnostic systems in healthcare.

In conclusion, we have developed and validated FOCUS, a system that transcends the role of a conventional diagnostic algorithm. It is a full-process, clinically-aligned framework—powered by a foundation model—engineered to automate the entire OCT diagnostic workflow. We have demonstrated that this end-to-end architecture achieves robust, expert-level performance. Crucially, by synthesizing slice-level data into a conclusive patient-level diagnosis, FOCUS directly solves the critical workflow and integration gaps that have hindered the clinical deployment of prior AI models.